\documentclass[11pt]{article}
\usepackage{setspace}
\usepackage[caption=false]{subfig} 
\usepackage{graphicx,caption}
\usepackage{authblk} 
\usepackage{indentfirst}
\usepackage[left=3cm,right=2cm,top=2.5cm,bottom=2cm]{geometry}
\usepackage{float}
\usepackage{array}
\usepackage{xcolor}
\usepackage{mathtools}
\usepackage{amsmath,amsfonts}
\usepackage{bbm}
\usepackage{xspace}
\usepackage{hyperref}
\usepackage{hyperref}
\usepackage[capitalise,noabbrev]{cleveref}
\crefname{appendixfigure}{Supplementary Figure}{SI Figures}
\usepackage{todonotes}
\usepackage[utf8]{inputenc}
\usepackage{siunitx}
\usepackage{bm}
\usepackage{float}

\usetikzlibrary{patterns}
\newcommand*{\TitleFont}{%
      \usefont{\encodingdefault}{\rmdefault}{b}{n}%
      \fontsize{16}{20}%
     \selectfont}
\usepackage[percent]{overpic}

\makeatletter
\renewcommand{\maketitle}{\bgroup\setlength{\parindent}{0pt}
\begin{flushleft}
 \fontsize{16}{32}
 \textbf{\@title}
 
   \fontsize{12}{24}
  \@author
\end{flushleft}\egroup
}

\newfam\bboardfam
\font\tenbboard=msbm10  
 \font\sevenbboard=msbm7
   \font\fivebboard=msbm5 
\textfont\bboardfam=\tenbboard
\scriptfont\bboardfam=\sevenbboard
\scriptscriptfont\bboardfam=\fivebboard

\title{\TitleFont Engineering flexible machine learning systems by traversing functionally-invariant paths}

\author
{Guruprasad Raghavan$^{1}$, Bahey Tharwat$^{2}$,Surya Narayanan Hari $^{1}$, Dhruvil Satani $^{1}$, \ \ \ \ Matt Thomson $^{1\ast}$\\
\normalsize{$^{1}$Department of Biology and Biological Engineering, Caltech}\\
\normalsize{1200 E California Blvd, Pasadena, CA 91125 }\\
\normalsize{$^{2}$Alexandria University}\\
\normalsize{Alexandria, Egypt}\\
\normalsize{$^\ast$To whom correspondence should be addressed; \\mthomson@caltech.edu\\ graghava@caltech.edu}
}

\date{}

\begin{document} 
\maketitle 
\subsection*{Abstract} 
\noindent \textbf{Transformers have emerged as the state of the art neural network architecture for natural language processing and computer vision. In the foundation model paradigm, large transformer models (BERT, GPT3/4, Bloom, ViT)  are pre-trained on self-supervised tasks such as word or image masking, and then, adapted through fine-tuning for downstream user applications including instruction following and Question Answering. While many approaches have been developed for model fine-tuning including low-rank weight update strategies (eg. LoRA), underlying mathematical principles that enable network adaptation without knowledge loss remain poorly understood.  Here, we introduce a differential geometry framework, functionally invariant paths (FIP), that provides flexible and continuous adaptation of neural networks for a range of machine learning goals and network sparsification objectives. We conceptualize the weight space of a neural network as a curved Riemannian manifold equipped with a metric tensor whose spectrum defines low rank subspaces in weight space that accommodate network adaptation without loss of prior knowledge. We formalize adaptation as movement along a geodesic path in weight space while searching for networks that accommodate secondary objectives. With modest computational resources, the FIP algorithm achieves comparable to state of the art performance on continual learning and sparsification tasks for language models (BERT), vision transformers (ViT, DeIT), and the CNNs. Broadly, we conceptualize a neural network as a mathematical object that can be iteratively transformed into distinct configurations by the path-sampling algorithm to define a sub-manifold of weight space that can be harnessed to achieve user goals.  }

\subsection*{Introduction}
Transformers are now the state of the art machine learning paradigm for natural language understanding,  computer vision, biological sequence analysis, and context sensitive reasoning tasks \cite{he2022masked,taori2023alpaca,bommasani2021opportunities,brown2020language}. As models have scaled in parameter number, models trained on generic masking tasks have exhibited emergent behaviors including zero-shot task performance,  generalized reasoning, and instruction following \cite{brown2020language,devlin2018bert,openai2023gpt4}.  
In the `foundation model’ paradigm \cite{bommasani2021opportunities}, transformers with $10^8$ to $10^{12}$ parameters are trained over large data sets on self-supervised tasks such as masked language modeling, causal language modeling, or image masking \cite{hoffmann2022empirical,openai2023gpt4,dosovitskiy2020image,devlin2018bert}.  Following self supervised training, models can be adapted to increase performance on specific applications including question/answer,  instruction following, distillation of financial or medical documents,  and sparsified or quantized to reduce memory requirements and inference speeds in deployment environments. 

Due to the central role of model adaptation for transformer optimization and deployment, many algorithms have emerged for updating model weights to increase performance without experiencing a catastrophic loss of the knowledge gained during self-supervised pre-training.  For example, LoRA (Low Rank Adaptation) exhibits impressive performance on fine-tuning of billion parameter models through low rank weight updates enforced by matrix factorization \cite{hu2021lora}. Empirical experiments with LoRA find that low-rank updates enable fine tuning of models with \text{100B} parameters in benchmark tasks.  However, the performance of fine-tuning and sparsification frameworks remains predominantly grounded in empirical results. The machine learning community would benefit from mathematical tools that provide general insights and unification of model adaptation strategies within a common framework. Besides LoRA, many different weight update strategies have been defined. Methods like orthogonal gradient descent (OGD) and elastic weight consolidation propose different criteria for updating weights in directions that do not impact performance on previously learned tasks. Yet, most current methods are based on local heuristics, for example, selecting gradient steps that are orthogonal to gradient steps taken for previously learned tasks. Sparsification frameworks execute prune/fine-train cycles to discover a core-sub network capable of executing the desired behavior. Mathematical tools that provide deeper insight into how the global, geometric structure of weight space enables or complicates adaptation might provide both conceptual principles and new algorithms. 

\begin{figure*}[p]
\vspace*{-1.5cm}
\centering
 \advance\topskip-1cm
    \includegraphics[scale= 1]{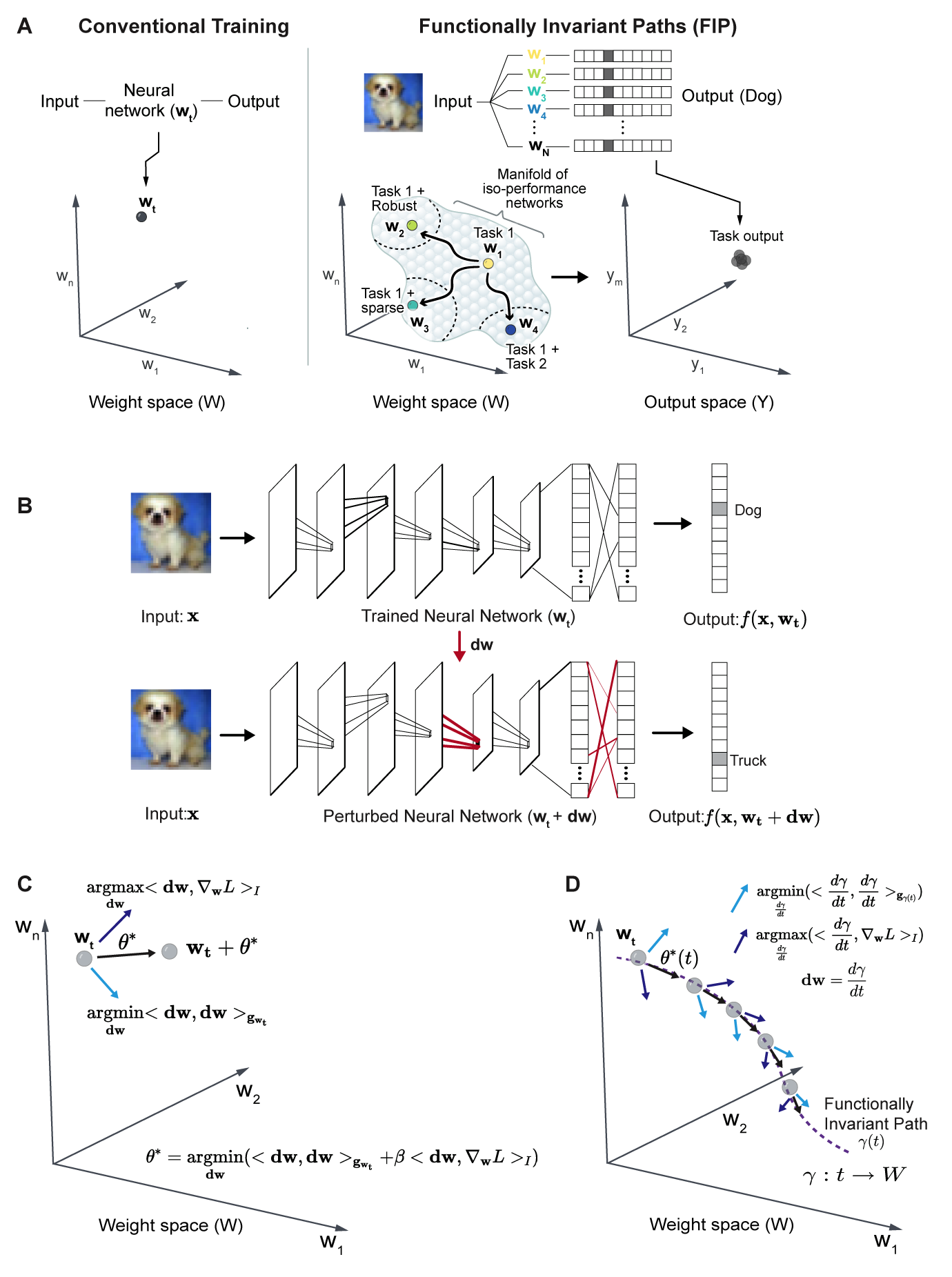}
    \caption{\textbf{Differential Geometric Framework for constructing functionally invariant paths (FIPs) in weight space.} (A) (Left) Conventional training on a task finds a single trained network ($\mathbf{w_t}$) solution. (Right) The FIP strategy discovers a submanifold of iso-performance networks ($\mathbf{w_1}$, $\mathbf{w_2}$, ...,$\mathbf{w_N}$) for a task of interest, enabling the efficient search for networks endowed with adversarial robustness ($\mathbf{w_2}$), sparse networks with high task performance ($\mathbf{w_3}$) and for learning multiple tasks without forgetting ($\mathbf{w_4}$). (B) (Top) A trained convolutional neural network with weight configuration ($\mathbf{w_t}$), represented by lines connecting different layers of the network, accepts an input image, $\mathbf{x}$, and produces a 10-element output vector, $f(\mathbf{x},\mathbf{w_t})$. (Below) Perturbation of network weights by $\mathbf{dw}$ results in a new network with weight configuration $\mathbf{w_t} + \mathbf{dw}$ with an altered output vector, $f(\mathbf{x},\mathbf{w_t}+\mathbf{dw})$, for the same input, \textbf{x}. 
    (C) The FIP algorithm identifies weight perturbations, $\mathbf{\theta}^*$ that minimize distance moved in output space while maximizing alignment with gradient of a secondary objective function ($\nabla_\mathbf{w} L$). The light-blue arrow indicates an $\epsilon$-norm weight perturbation that minimizes distance moved in output space, dark-blue arrow is an $\epsilon$-norm weight perturbation that maximizes alignment with gradient of objective function, $L(\mathbf{x}, \mathbf{w})$. The secondary objective function $L(\mathbf{x}, \mathbf{w})$ is varied to solve distinct machine learning challenges. (D) The path sampling algorithm defines functionally invariant paths, $\gamma(t)$, through iterative identification of $\epsilon$-norm perturbations ($\mathbf{\theta}^*(t)$) in the weight space.
    \label{fig:fig1}}
\end{figure*}

In artificial neural networks, network function is encoded in the mathematical weights that determine the strength of connections between neural units (Fig. \ref{fig:fig1}A,B). Machine learning procedures train multi-layered neural networks to solve problems by adjusting the weights of a network based on an objective function that encodes the performance of a network on a specific task. Standard learning methods, like back-propagation and gradient descent \cite{rumelhart1986learning}, adjust network weights to define a single, optimal weight configuration to maximize performance on a task specific objective function using training data. Training the network on new tasks through the traditional paradigm of stepping along the gradient of the task-specific objective function adjusts the networks' weights, inevitably resulting in the loss of information from previous tasks. 

Unlike contemporary artificial neural nets,  neural networks in the human brain perform multiple functions and can flexibly switch between different functional configurations based on context, goals or memory \cite{minxha2020flexible}. Neural networks in the brain are hypothesized to overcome the limitations of a single, optimal weight configuration and perform flexible tasks by continuously `drifting' their neural firing states and neural weight configurations, effectively generating large ensembles of degenerate networks \cite{mau2020brain, stringer2019spontaneous, masset2022drifting,geva2023time,driscoll2022representational}.Fluctuations might enable flexibility in biological systems by allowing neural networks to explore a series of network configurations while responding to sensory input.

Here, we develop a geometric framework and algorithm to construct path connected sets of neural networks that solve a given machine learning task. Conceptually, we consider path-connected sets of neural networks, rather than single-networks (isolated points in weight space) to be the central objects of study and application. By building sets of networks rather than single networks, we search within a sub-manifold of weight space for networks that solve a given machine learning problem while accommodating a broad range of secondary goals.Historically, results in theoretical machine learning and information geometry have pointed to the geometry of a model’s loss landscape as a potential resource for model adaptation. An emergent property of over-parameterized models is the existence of parameter degeneracy where multiple settings of the model parameters can achieve identical performance on a given task. Geometrically, parameter degeneracy leads \cite{machta2013parameter} to `flat' objective functions \cite{hochreiter1997flat,hochreiter1994simplifying,tsuzuku2020normalized} along which network weights can be modified without loss of performance. To discover invariant subspaces, we introduce a Riemmanian metric into weight space, a tensor,  that measures, at every point in parameter space, the change in model output given an infinitesimal movement in model parameters. The metric provides a mathematical tool for identifying low-dimensional subspaces in weight space where parameter changes have little impact on how a neural network transforms input data. Riemannian metrics are widely used in physical theories to study the dynamics of particles on curved manifolds and space-times. Geometrically, the metric discovers flat directions in weight space along which we can translate a neural network without changing functional performance. 

Using the Riemannian weight space metric, we develop an algorithm that constructs functionally invariant paths in weight space that maintain network performance while ‘searching-out’ for other networks that satisfy additional objectives. The algorithm identifies long range paths in weight space that can integrate new functionality without information loss. We apply the functionally invariant paths, FIP, framework to natural language (BERT) and vision transformers (ViT, DeIT) as well as CNN architectures. Our framework generates results that are in line with state of the art performance for adaptation and sparsification tasks on academic grade hardware. Our approach provides mathematical machinery that yields insight into how low-dimensional geometric structure can be harnessed for model adaptation without information loss. More broadly, we consider language models as objects acted upon by transformations in series and we show that the transformations of models are intrinsically non-abelian. The unified framework provides a general attack on a range of model adaptation problems and reveals connections between the mathematical theory of differential geometry and the emergent properties of large language and vision models. 

\subsection*{Riemannian metric enables construction of functionally invariant paths in weight space} 
We develop a mathematical framework that allows us to define and explore path-connected sets of neural networks that have divergent weight values but similar output on training data. We view the weight-space of a neural network as a Riemannian manifold equipped with a local distance metric \cite{amari2016information, benn1987introduction}. Using differential geometry, we construct paths through weight space that maintain the functional performance of a neural network while adjusting network weights to flow along a secondary goal (Fig 1A). The secondary goal can be general, so that the framework can be applied to train networks on new classification tasks, sparsify networks, and mitigate adversarial fragility. 

The defining feature of a Riemannian manifold is the existence of a local distance metric. We construct a distance metric in weight space that defines the distance between two nearby networks to be their difference in output. We consider a neural network to be a smooth function, $f(\mathbf{x};\mathbf{w})$, that maps an input vector, $\mathbf{x} \in \mathbb{R}^{\text{k}}$, to an output vector, $f(\mathbf{x};\mathbf{w})  = \mathbf{y}\in \mathbb{R}^{\text{m}}$,  where the map is parameterized by a vector of weights, $\mathbf{w} \in \mathbb{R}^\text{n}$, that are typically set in training to solve a specific task. We refer to $W= \mathbb{R}^n$ as the $\textit{weight space}$ of the network, and we refer to $\mathcal{Y} = \mathbb{R}^m$  as the $\textit{output space}$ as shown in Fig. \ref{fig:fig1}B,C \cite{mache2006trends}. For pedagogical purposes, we will consider the action of $f$ on a single input, $\mathbf{x}$. In the supplement we show that our results extend naturally to an arbitrary number of inputs {$\mathbf{x_i}$} (Supporting Information). 

We initially ask how the output, $f(\mathbf{x};\mathbf{w})$, of a given neural network changes for small changes in network weights.  Given a neural network with weights $\mathbf{w_t}$,  a fixed input $\mathbf{x}$,  we can compute the output of the perturbed network, $\mathbf{w_t} + \mathbf{dw}$  for an infinitesimal weight perturbation, $\mathbf{dw}$ as
\begin{align}
f(\mathbf{x},\mathbf{w_t} +\mathbf{dw}) \approx f(\mathbf{x},\mathbf{w_t}) +\mathbf{ J_{w_t}} \ \mathbf{dw},
\vspace{-2mm}
\end{align}
where $\mathbf{ J_{w_t}}$ is the Jacobian of $f(\mathbf{x},\mathbf{w_t})$ for a fixed $\mathbf{x}$, $J_{i j} = \frac{\partial f_i}{\partial w_j}$, evaluated at $\mathbf{w_t}$. 

Thus, the total change in network output for a given weight perturbation $\mathbf{dw}$ is
\begin{align}
\label{local metric}
|f(\mathbf{x},\mathbf{w_t}+\mathbf{dw})-f(\mathbf{x},\mathbf{w_t})|^2  &= \mathbf{dw}^T \ (\mathbf{J_{w_t}(\mathbf{x})}^T \
\mathbf{J_{w_t}(\mathbf{x})})  \  \mathbf{dw} \\ |\langle\mathbf{dw},\mathbf{dw}\rangle_{\mathbf{g_{w_t}}}|^2 &= 
 \mathbf{dw}^T \ \mathbf{g_{w_t}(\mathbf{x})} \ \mathbf{dw} \nonumber
\end{align}
where $\mathbf{g_{w_t}(\mathbf{x})}$ = $\mathbf{J_{w_t}}$($\mathbf{x})^T$ $\mathbf{J_{w_t}(\mathbf{x})}$ is the metric tensor evaluated at the point $\mathbf{w_t} \in W$ for a single data point, $\mathbf{x}$. The metric tensor is an $n \times n$ symmetric matrix that allows us to compute the change in network output given movement of the network along any direction in weight space as $\langle \mathbf{dw}, \mathbf{dw} \rangle_{\mathbf{g_{w_t}(x)}}$. As we move through the weight space, $W$, the metric tensor continuously changes allowing us to compute the infinitesimal change in network output while moving along a path $\mathbf{\gamma(t)}$ as the tangent vector $\psi(t) = \frac{\text{d} \gamma(t)}{\text{d}t}$. 

At every point in weight space, the metric allows us to discover directions $\mathbf{dw}$ of movement that have large or small impact on the output of a network.  As we move along a path, $\gamma(t) \subset W$, in weight space, we sample a series of neural networks over time, $t$. 
Using the metric we can define a notion of `output velocity', $\mathbf{v} = \frac{\text{d} f(\mathbf{x}, \gamma(t))}{\text{dt}}$, that quantifies the distance a network moves in output space for each local movement along the weight space path $\gamma(t)$. We seek to identify `Functionally invariant paths (FIPs)’ in weight space along which the output velocity is minimized for a fixed magnitude change in weight. To do so, we solve the following  optimization problem

\begin{align}
    \psi^*(t) &= \mathop{\mathrm{argmin}}_\frac{d\gamma}{dt}  \langle \frac{d\gamma}{dt},  \frac{d\gamma}{dt}\rangle_\mathbf{g_
    {\gamma(t)}} \hspace{5mm} \nonumber \\ &\ \text{s. t. } \ \langle\frac{d\gamma}{dt}, \frac{d\gamma}{dt}\rangle_I = \epsilon 
    \label{eqn-FIP}
\end{align}

where we attempt to find a direction $\psi^*(t)$ along which to perturb the network, such that it is $\epsilon$ units away from the base network in the weight space  (in the euclidean sense, $\langle\frac{d\gamma}{dt}, \frac{d\gamma}{dt}\rangle_I$ = $\epsilon$) while minimizing the distance moved in the networks' output space, given by $\langle\frac{d\gamma}{dt},  \frac{d\gamma}{dt}\rangle_\mathbf{g_{\gamma(t)}}$. Here, $I$ is an identity matrix, with the inner product $\langle\frac{d\gamma}{dt}, \frac{d\gamma}{dt}\rangle_I$ capturing the euclidean distance in the weight space \cite{weisstein2014metric}. The optimization problem is a quadratic program at each point in weight space. The metric $\mathbf{g}$ is a matrix that takes on a specific value at each point in weight space, and we aim to identify vectors $\mathbf{\psi}^*(t) = \frac{d\gamma(t)}{dt}$, that minimize the change in functional output of the network. 

We will often amend the optimization problem with a second objective function $L(\mathbf{x}, \mathbf{w})$.  We can enumerate paths that minimize the functional velocity in the output space while moving along the gradient of the second objective ($\nabla_\mathbf{w} L$). We define a path-finding algorithm that identifies a direction $\psi^*(t)$ in weight space by minimizing the functional velocity in the output space while moving along the gradient of the second objective ($\nabla_\mathbf{w} L$)

\begin{align}
    \psi^*(t) &= \mathop{\mathrm{argmin}}_\frac{d\gamma}{dt} (\langle \frac{d\gamma}{dt}, \frac{d\gamma}{dt} \rangle_{\mathbf{g}_{\gamma(t)}}+\beta \langle\frac{d\gamma}{dt}, \nabla_\mathbf{w} L\rangle_I) \nonumber \\ 
    &\text{s. t. }  \langle\frac{d\gamma}{dt}, \frac{d\gamma}{dt}\rangle_I = \epsilon \label{eqn-dirFIP}  
\end{align}
    

where the first term, $\langle \frac{d\gamma}{dt}, \frac{d\gamma}{dt} \rangle_{\mathbf{g}_{\gamma(t)}}$,  identifies functionally invariant directions while the second term, $\langle\frac{d\gamma}{dt}, \nabla_\mathbf{w} L\rangle_I$, biases the direction of motion along the gradient of a second objective and $\beta$ weighs the relative contribution of the two terms. When $L$ = 0, the algorithm merely constructs paths in weight space that are approximately isofunctional ($\theta^*(t) = \psi^*(t)$), i.e. the path is generated by steps in the weight space comprising of networks with different weight configurations while preserving the input-output map. $L(\mathbf{x}, \mathbf{w})$ can also represent the loss function of a second task, for example a second input classification problem. In this case, we identify vectors that simultaneously maintain performance on an existing task (via term 1) while also improving performance on a second task by moving along the negative gradient of the second task loss function, $\nabla_\mathbf{w} L$. We think of constructing FIPs with different objective functions ($L(\mathbf{x}, \mathbf{w})$) similar to applying different ``operations" to neural networks that identify sub-manifolds in the weight space of the network that accomplish distinct tasks of interest. 

To approximate the solution to Eq-\ref{eqn-dirFIP}, in large neural networks, we developed a numerical strategy that samples points in an $\epsilon$ ball around a given weight configuration, and then performs gradient descent to identify vectors $\mathbf{\theta}^*(t)$. In the appendix, we extend the metric formulation to cases where we  consider a set of N training data points, $\mathbf{X}$, and view $\mathbf{g}$ as the average of metrics derived from individual training examples. $ \mathbf{g_w} = \mathbf{g_w(X)} = \Sigma_{i=1}^N\mathbf{g_w(x_i)}/N$. The metric, $\mathbf{g}$, provides a local measure of \textit{output distance} on the Riemannian manifold $(W,\mathbf{g_{\mathbf{w}}})$. At each point in weight space, the metric defines the length, $\langle \mathbf{dw}, \mathbf{dw} \rangle_{\mathbf{g}_{\mathbf{w}}}$, of a local perturbation by its impact on the functional output of the network (Fig. \ref{fig:fig1}B,C). 

We note that the mathematical framework provides avenues for immediate generalization. On any Riemannian manifold, we can define geodesic paths emanating from a point $x_p$ as paths of constant velocity $\langle \frac{d\gamma}{dt}, \frac{d\gamma}{dt} \rangle_{\mathbf{g}_{\gamma(t)}} = v_0$ (Supporting Materials) such geodesic paths have a constant, potentially non-zero, rate of performance decay on task 1 during adaptation. 

\begin{figure*}[p]
\vspace*{-1.5cm}
\centering
 \advance\topskip-1cm
\includegraphics[scale=1]{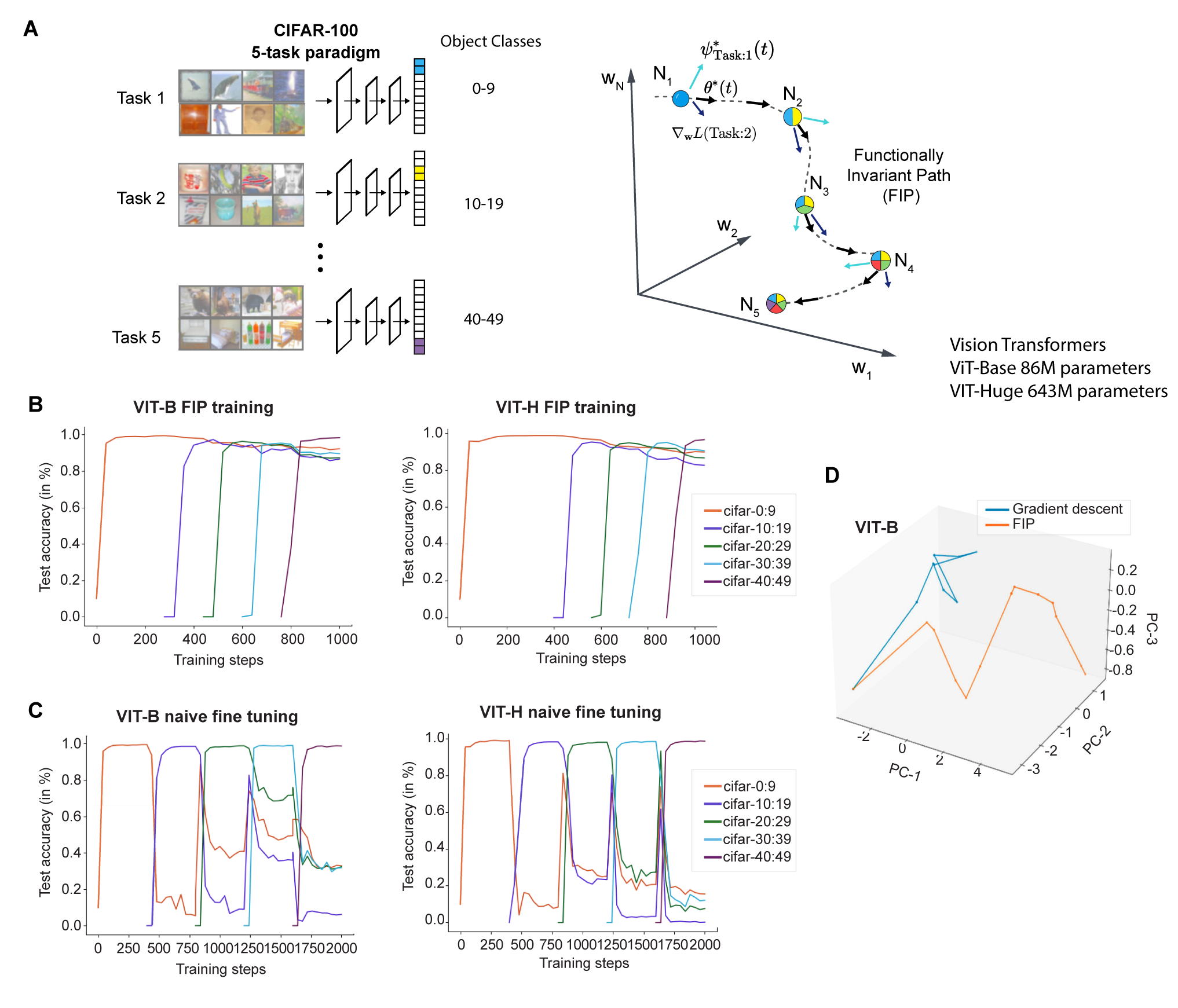} \caption{\textbf{Vision Transformers learn sequential tasks without catastrophic forgetting by traversing FIPs.} (A)Five task continual learning paradigms where each task is a 10 way object classification, where the classes are taken from CIFAR100. (ii) Schematic of FIP construction in weight space to train networks on 5 tasks sequentially using the ViT (Vision Transformer) Base and Huge. (B,C) Test accuracy for ViT-B and ViT-H using FIP or naive fine-tuning.  Following continual training, FIP achieves  $91.2\%$ and $89.3\%$ test accuracy using ViT-B and Vit-H respectively for all five tasks. Baseline performance for ViT-B trained on all five tasks simultaneously is $94.5\%$. Training for Vit-B using an NVIDIA RTX2060 6GB machine took $\approx 3.5 \text{hours}$ for each sub-task with the FIP and $\approx 2.5 \text{hours}$ with fine-tuning. Training for Vit-H using an NVIDIA RTX3090 24GB machine took $\approx 4.8 \text{hours}$ for each sub-task with the FIP and $\approx 3.9 \text{hours}$ with fine-tuning. (D) PCA plots of the FIP (orange) and gradient descent fine tuning (blue) weight space path showing that the FIP allows long range exploration of weight space.}
\label{fig:fig2}
\end{figure*}



\subsection*{Application of the functionally invariant path (FIP) framework to continual learning, sparsification, and adversarial robustness tasks}

\subsection*{FIP enables continual learning with ViT vision transformers }
The FIP framework allows us to address a series of model adaptation goals within a common geometric framework. To demonstrate continual learning, we applied the FIP to adapt the ViT vision image transformer and BERT language model in continual learning without catastrophic forgetting.  We train a neural network on a base task and modulate the weights in the network to accommodate additional tasks by solving the optimization problem in Eq-\ref{eqn-dirFIP},  setting $L(\mathbf{x}, \mathbf{w})$ as the classification loss function specified by the additional task while $\langle \frac{d\gamma}{dt}, \frac{d\gamma}{dt}\rangle_{\mathbf{g}_{\text{Task}1}}$ measures distance moved in the networks' output using the metric from the initial task (in Eq-\ref{eqn-dirFIP}A. To accommodate the additional tasks, we append output nodes to the base network and solve the optimization problem for a fixed value of $\beta$ by simultaneously minimizing the distance moved in the networks' output space (light-blue arrow) corresponding to the first task while maximizing alignment with the gradient of $L(\mathbf{x}, \mathbf{w})$ encoding the classification loss from the second task. In this manner, we construct a Functionally invariant path (FIP) (purple dotted line) in weight space generating a network that performs both Task 1 and Task 2. 

We used a standard continual learning task \cite{kaushik2021understanding} (Fig \ref{fig:fig2}A). We split the CIFAR-100 data set into a series of sub-tasks where each sub-task requires the network to identify $10$ object categories (Fig \ref{fig:fig2}A). Previously, the state of the art performance for this task was achieved by the ResNet CNN \cite{kaushik2021understanding}. ViT networks can potentially realize vast performance gains over ResNet architectures as the baseline performance for ResNet is $\approx 80 \% $ accuracy, and we observed that ViT exhibits $94.5 \%$ accuracy when fine-tuned on CIFAR100 \cite{dosovitskiy2020image},\cite{van2020brain} generative replay achieving CIFAR-100 continual learning accuracy of $\approx 80 \%$. 

We applied the FIP algorithm to achieve continual learning on SplitCIFAR using both the ViT-B (86M parameter) and ViT-H (632M parameter) architectures. In each case, a single task requires the network to learn to recognize $10$ CIFAR classes where we used $5000$ images (of 6000 total) per class with batch size $512$. We used a single NVIDIA RTX2060 6GB for VitB and RTX3090 24GB for ViT-H (Fig \ref{fig:fig2})B. After continual training 5 SpliTCIFAR sub-tasks, ViT-B achieved a mean performance of $ 91.2\%$ and ViT-H $89.3 \%$ compared with $94.5 \%$ performance for ViT-B trained on all $50$ CIFAR classes simultaneously without continual learning (Fig \ref{fig:fig2}B,C). Training for Vit-B using an NVIDIA RTX2060 6GB machine took $\approx 3.5 \ \text{hours}$ for each sub-task with the FIP and $\approx 2.5 \ \text{hours}$ with fine-tuning. Training for Vit-H using an NVIDIA RTX3090 24GB machine took $\approx 4.8 \ \text{hours}$ for each sub-task with the FIP and $\approx 3.9 \ \text{hours}$ with standard fine-tuning. 

Thus, the FIP procedure enables VIT-B and Vit-H to learn new tasks without information loss while achieving a significantly higher than continual learning methods that have been conventionally applied to the ResNet network \cite{van2020brain}. We note that Model Zoo, a highly innovative framework that grows neural networks, has high performance $\approx 95 \%$ in the SplitCIFAR task \cite{ramesh2021model}. The strength of our method is that it can be applied to any existing transformer or CNN architecture. We have now applied out method to adapt large vision transformers and can achieve performance below but in line with model zoo. 

\subsection*{FIP enables continual learning with BERT NLP transformer}

\begin{figure*}[p]
\vspace*{-1.5cm}
\centering
 \advance\topskip-1cm
\includegraphics[scale=1.0]{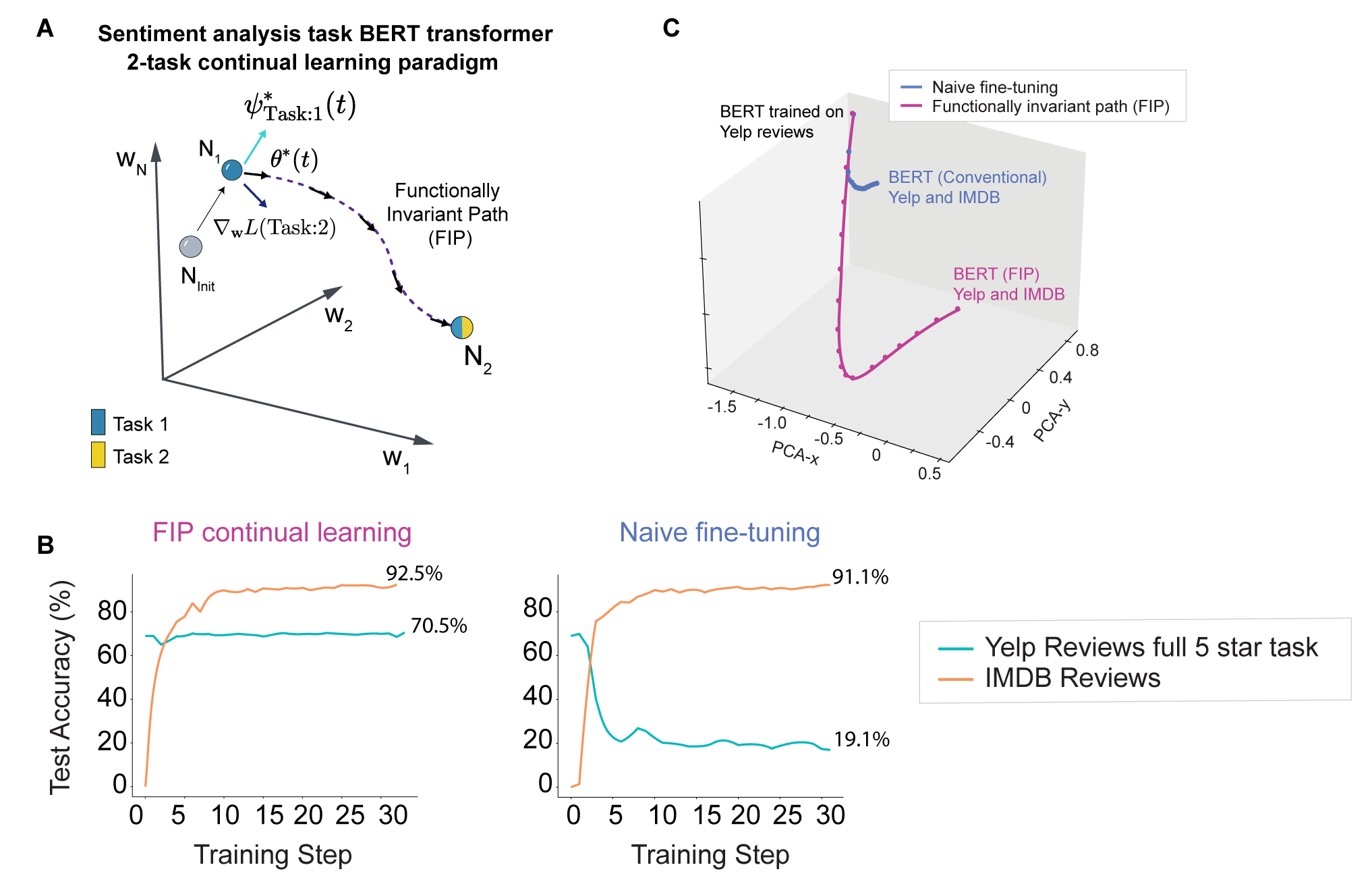} \caption{\textbf{The BERT transformer learns sequential sentiment analysis tasks without catastrophic forgetting by traversing a FIP.} (A)Schematic of FIP construction in weight space to train networks on 2 tasks sequentially. We initially train the BERT transformer on the Yelp Five star review prediction task and then add the IMDB sentiment analysis task.  (B) Test accuracy for BERT trained through continual learning via the FIP or with conventional fine-tuning. (C) Weight space paths for conventional fine tuning and FIP based training.}
\label{fig:fig2bert}
\end{figure*}

Next, we demonstrated the flexibility of the FIP approach by using the method to fine-tine the BERT network on the IMDB sentiment analysis task following initial training on Yelp full-five star review prediction task (Fig \ref{fig:fig2bert}). The BERT network has a total of twelve layers, or transformer blocks, with 12 self-attention heads in each layer and a total of 110 million parameters \cite{devlin2018bert}. Training BERT to detect customer opinions of a product based on text reviews left on websites like Yelp or IMDB, results in catastrophic forgetting, especially while sequentially training on multiple user-review datasets (say, yelp-reviews followed by IMDB) (Fig \ref{fig:fig2bert}A). The FIP maintains BERT performance on Yelp-reviews (at 70$\%$, blue) while increasing its accuracy on IMDB review classification (from 0$\%$ to 92$\%$, orange) (Fig \ref{fig:fig2bert}B). The FIP in BERT weight space (Fig. \ref{fig:fig2bert} ) is much longer than the route taken by conventional training, enabling global exploration of the BERT weight-landscape to identify networks that simultaneously maintain performance on Yelp-reviews while learning the IMDB sentiment classification. Conventional fine-tuning of BERT on IMDB reviews increases its performance on sentiment classification on IMDB (from 0$\%$ to 92$\%$, orange) while abruptly forgetting sentiment analysis on Yelp-reviews (dropping from an accuracy of 69.9$\%$ to 17$\%$, blue) within 30 training steps (Fig \ref{fig:fig2bert}B).

\begin{figure*}[p]
\vspace*{-1.5cm}
\centering
 \advance\topskip-1cm
\includegraphics[scale=.6]{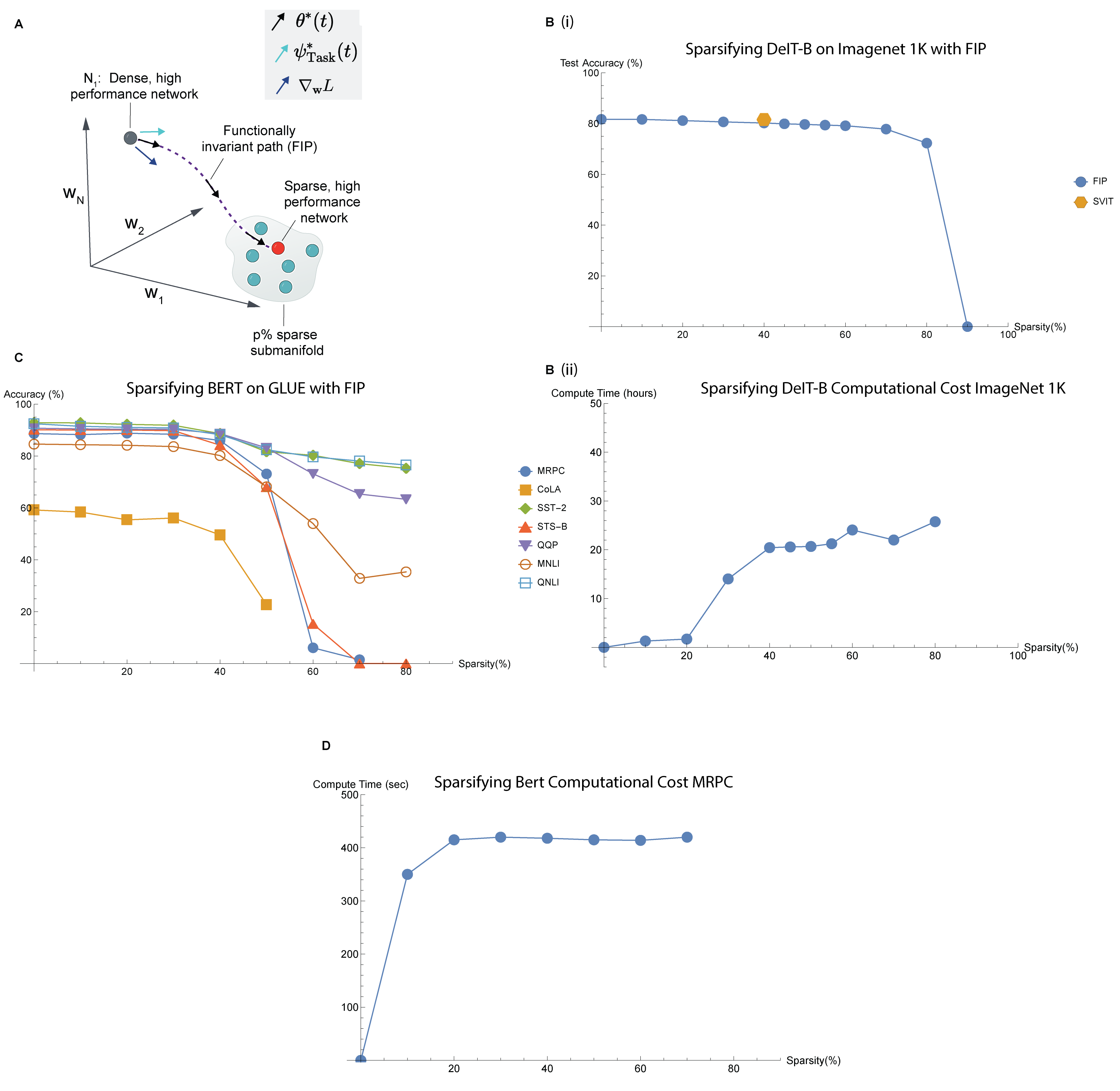} \caption{\textbf{Sparsification of vision and language transformers with FIP} (A) We applied the FIP algorithm to generate sparse versions (network weights set to 0) of the DeIT-B vision transformer performing the ImageNet1K image classification task. Using the FIP we generated networks with sparsity (fraction of weights set to 0) ranging from 0 to 80\%. We compared network performance to sparsification reported for DeIT-B using a state of the art method reported at NeurIPS in 2021 \cite{chen2021chasing}. FIP was able to achieve very minimal reductions in performance until $\approx 80 \%$ sparsity, but further attempts at parameter reduction failed (B) Compute time in hours for DeIT sparsification. (C) Sparsification of BERT on the nine GLUE natural language processing tasks. Sparsification performance is task dependent. and compute time (D) in seconds for the MRPC task on an Nvidia A100 machine donated by PaperSpace } 
\label{fig:figsparse}
\end{figure*}

\subsection*{Neural network sparsification with FIP algorithm}

The critical aspects of the FIP framework is that the framework generalizes and addresses a broad range of machine learning meta-problems by considering a more general set of secondary objective functions. In particular, we next apply the FIP framework to perform sparsification, reducing the number of non-zero weights, which is important for reducing the memory and computational footprint of a network \cite{blalock2020state}. To sparsify neural networks, we solve Eq-\ref{eqn-dirFIP}, the core FIP optimization problem, with a secondary loss function $L(w_t,w,p)$ that measures the euclidean distance between a network and it's p-sparse projection obtained by setting $p\%$ of the networks’ weights to zero (Fig \ref{fig:figsparse}A). 

Using the framework, we sparsified the vision transformer DeIT which has been used for benchmarking sparsification methods \cite{chen2021chasing} on vision transformers.  The paradigm uses the ImageNet1K data set, and attempts to sparsify DeIT. DeIT-base is an $86$ M parameter transformer model that was derived from ViT \cite{touvron2021training}.  We use the FIP algorithm to set weight parameters to zero without loss of performance on the Imagenet 1000 object image classification task. The simplicity of the FIP algorithm allowed us to achieve an entire range of target sparsities ranging from $0\%$ to $80\%$ sparsity. We found that FIP had performance very near that of the SViT network at the benchmark of $40\%$ sparsity with FIP performing at $80.22 \%$ and SViT performing at $81.56 \%$ accuracy (Fig \ref{fig:figsparse}B) (i) with compute times given (Fig \ref{fig:figsparse}B(ii)) on an Nvidia RTX3090 24GB. 

We, then, applied FIP to sparsify BERT base from $0 \%$ to $80 \%$ sparsity for all the GLUE natural language processing tasks. For this task, we obtained an Nvidia A100 machine from PaperSpace. The GLUE benchmark consists of three categories of natural language understanding tasks: (a) single-sentence tasks (CoLA and SST-2); (b) similarity and para phrase tasks (MRPC, QQP and STS-B); (c) inference tasks (MNLI, QNLI and RTE).  Again, because of its ease of use and efficiency, we were able to span the entire range of sparsities identifying differential performance across GLUE tasks (Fig \ref{fig:figsparse}C). FIP was able to generate sparse versions of BERT that had $81.65\%$ accuracy at $50\%$ sparsity on the SST-2 task. We provide compute times in seconds for sparsification of BERT on MRPC which runs efficiently on an Nvidia A100 machine (Fig \ref{fig:figsparse}D).

\subsection*{Path-connected sets of networks confer robustness against adversarial attack}

\begin{figure*}[p]
    \vspace*{-1.5cm}
    \centering
     \advance\topskip-1cm
    \includegraphics[width=\textwidth]{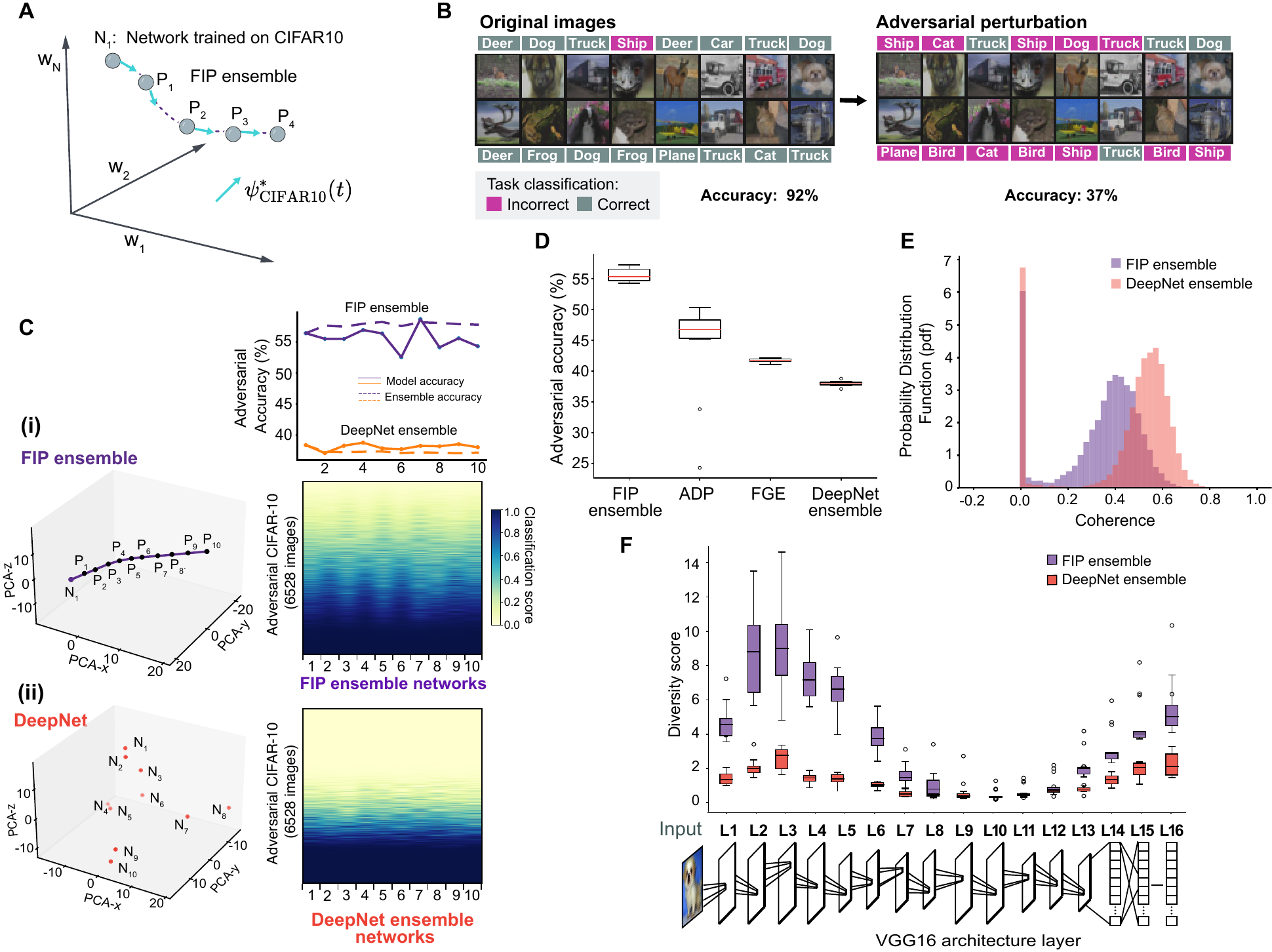} \caption{\textbf{FIPs in weight space generate ensembles of networks that confer adversarial robustness} (A) Schematic to generate FIP ensemble (P$_1$,...,P$_4$) by sampling networks along FIP (purple dotted line) beginning at network-N$_1$. FIP is constructed by identifying a series of weight perturbations that minimize the distance moved in networks' output space. 
    (B) Original CIFAR10 images (left) and Adversarial CIFAR-10 images (right) are shown. The text-labels (left, right) above the images are predictions made by a network trained on CIFAR-10. Trained networks' accuracy on the original and adversarial images are shown below. (C) (Top) Line-plot (solid) shows the individual network performance on adversarial inputs, (dashed) shows the joint ensemble accuracy on adversarial inputs, for FIP ensemble (purple) and DeepNet ensemble (orange). (i,ii-Left) FIP ensemble in purple (P$_1$,P$_2$,...,P$_{10}$) and DeepNet ensemble in orange (N$_1$,N$_2$,...,N$_{10}$) are visualized on weight space PCA. (i,ii-Right) Heatmaps depict classification score of networks in FIP ensemble and DeepNet ensemble on 6528 adversarial CIFAR-10 examples. (D) Boxplot compares adversarial accuracy (over 10k adversarial examples) across different ensembling techniques (n=3 trials). (E) Histogram of coherence values for FIP (purple) and DeepNet ensemble (orange).  (F) Boxplot shows the ensemble diversity score across VGG16 layers over n=1000 CIFAR10 image inputs. The cartoon below depicts the VGG16 network architecture. }
    \label{fig:fig4}
\end{figure*}

The path connected sets of networks generated by the FIP can also be applied to perform inference and increase the robustness of inference tasks to data perturbation.  Although deep CNN networks have achieved remarkable performance on image-recognition tasks,  human-imperceptible additive perturbations, known as adversarial attacks, can be applied to an input image and induce catastrophic errors in deep neural networks (Fig. \ref{fig:fig4}B). The FIP algorithm provides an efficient strategy to increase network robustness and mitigate adversarial failure by generating path-connected sets of networks with diverse weights. We then apply the path connected network sets to perform robust image classification by averaging their output. 

To demonstrate that the FIP algorithm can mitigate adversarial attacks, we trained a 16 layered CNN, VGG16 with 130 million parameters to classify CIFAR10 images with $92 \%$ test accuracy.  We, then, generated adversarial test images using the projected gradient descent attack strategy. On adversarial test images, the performance of VGG16 dropped to $37\%$ (Fig. \ref{fig:fig4}B, Supplementary). To mitigate the adversarial performance loss, we applied the FIP algorithm to generate an ensemble of functionally invariant networks by setting $L=0$ in the optimization problem in  Eq-\ref{eqn-dirFIP} and setting  $\langle\frac{d\gamma}{dt},\frac{d\gamma}{dt}\rangle_{\mathbf{g}_{\text{CIFAR10}}}$ to be the distance moved in the networks' output space for CIFAR-10 images. We use the FIP ensemble to classify images by summing the `softmaxed' outputs of the ensemble. 

Using an ensemble of ten networks sampled along an FIP, we achieve accuracy of $55.61$ $\pm$ 1.1 $\%$ surpassing the performance of the DeepNet ensemble performance (composed of 10 independently trained deep networks) by 20.62$\%$ (Fig. \ref{fig:fig4}C). The FIP ensemble’s adversarial performance also surpasses other state of the art ensemble approaches including Adaptive Diversity Promoting (ADP, $43.84 \pm 7.8 \%$) ensemble and the Fast Geometric Ensembling (FGE, $41.7 \pm 0.34$) method. The two factors contributing to the FIP ensemble's robustness are (i) high intra-ensemble weight diversity, calculated by the representation diversity score (Supplementary) and (ii) low coherence (Supplementary) with a trained surrogate network (used to generate adversarial images) (Fig. \ref{fig:fig4}E,F). FIP Networks have a higher representation diversity score in their early processing layers, from Layer 1 to layer 6, when compared to the DeepNet ensemble, indicating that individual networks in the FIP ensemble extract different sets of local features from the adversarial image, preventing networks from relying on similar spurious correlations for image classification. We speculate that weight/parameter diversity in the FIP ensemble leads to differential susceptibility to adversarial examples but consistent performance on training and test examples.

\subsection*{Generating path-connected sets of language models through iterative traversal of weight space paths}

 \begin{figure*}[p]
    \vspace*{-1.5cm}
    \centering
     \advance\topskip-1cm
    \includegraphics[width = \textwidth]{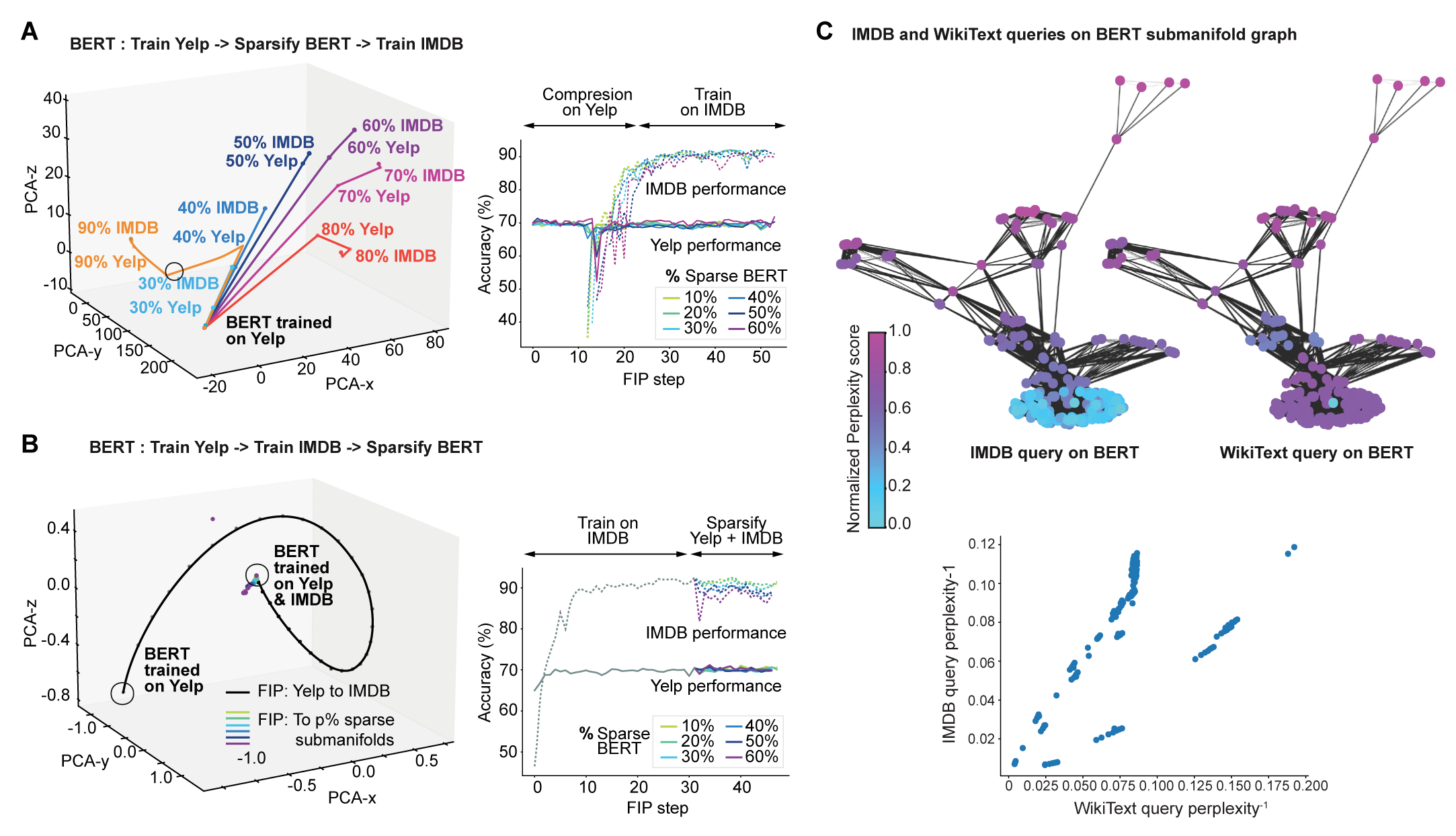} \caption{\textbf{Non-abelian, iterative transformation of BERT by FIP composition}  (A, left) FIP initially identify high performance sparse BERTs (for sparsities ranging from 10$\%$ to 80$\%$) followed by re-training on IMDB. (A, right) BERT accuracy on Yelp (solid) and IMDB (dashed) dataset along the FIP. (B, left) FIP initially retrains BERT on new task (IMDB) and then discovers a range of sparse BERTs. (B, right) BERT accuracy on Yelp (solid) and IMDB (dashed) dataset along the FIP. (C) Graph connected set of 300 BERT models trained on sentence completion on wikipedia and yelp datasets, colored by perplexity scores evaluated using two new query datasets, wikitext and imdb. Nodes correspond to individual BERT models and edges correspond to euclidean distance between BERT models in weight space. (C, rightmost) Scatter plot of inverse perplexity scores for two queries - IMDB and WikiText datasets.}
    \label{fig:fig5}
\end{figure*}

Conceptually,  the most important aspect of the FIP framework is that it unifies a series of machine learning meta-tasks (continual learning, sparsification) into a single mathematical framework. Mathematically, when we solve Equation \ref{eqn-dirFIP} with a given secondary loss function, we move an existing network $\textbf{w(0)}$ along a path in weight space generating a new network $\textbf{w(t)}$ where the parameter $t$ increments the length of a path. Each additional loss function generates a new transformation of a base network, for example, generating a network adjusted to accommodate an additional data analysis problem or a secondary objective like sparsification. Network transformation maps  can be iterated and applied to generate a family of neural networks optimized for distinct sub-tasks. The resulting path connected set of neural networks can then be queried for networks that achieve specific user goals or solve additional machine learning problems. 

The problem of customization is particularly important for transformer networks \cite{vaswani2017attention,brown2020language,liu2019roberta,wolf2020transformers}. Transformers like BERT and Roberta are typically trained on generic text corpus like the Common Crawl, and specialization of networks on specific domains is a major goal \cite{bommasani2021opportunities,hu2021lora}. We  applied the iterative FIP framework to generate a large number of natural language processing networks customized for different sub-tasks and sparsification goals .  Transformer networks incorporate layers of attention heads that provide contextual weight for positions in a sentence.  Transformer networks are often trained on a generic language processing task like sentence completion where the network must infer missing or masked words in a sentence. Models are, then, fine tuned for specific problems including sentiment analysis, question answering, text generation, and general language understanding \cite{vaswani2017attention}. Transformer networks are large containing hundreds of millions of weights, and so model customization can be computationally intensive. 

We applied the FIP framework to perform a series of model customization tasks through iterative application of FIP transformations on distinct goals. The BERT network has a total of twelve layers, or transformer blocks, with 12 self-attention heads in each layer and a total of 110 million parameters \cite{devlin2018bert}. Using the FIP framework, we applied  the two operations (Continual Learning (CL) (Fig \ref{Fig fig:fig2bert}), Compression (Co) (Fig \ref{fig:figsparse}) sequentially to BERT models trained on a range of language tasks, by constructing FIPs in the BERT weight space using different objective functions ($L(\mathbf{x}, \mathbf{w})$) while solving the optimization problem in Eq-\ref{eqn-dirFIP}. 

We demonstrated that we could apply the operations CL and Co in sequence. Beginning with the BERT base model, we applied the FIP model to generate networks that could perform YELP and IMDB sentiment analysis, and then compressed the resulting networks to generate $6$ distinct networks with sparsity ranging from $10-60 \%$ where all sparsified networks maintained performance on the sentiment analysis tasks (Fig. \ref{fig:fig5}A). We, then, performed the same operations but changed the order of operations (Fig. \ref{fig:fig5}B).  The models generated through application of CL Co(w) and Co CL(w) achieved similar functional performance in terms of sparsification and task performance but had distinct weight configurations. 

In total, we used the FIP framework to generate networks using a set of different natural language tasks and compression tasks yielding a path-connected set of 300 BERT models (Fig \ref{fig:fig5} C ). The $300$ networks define a sub-manifold of weight space that contains models customized for distinct sub-tasks. The sub-manifold, then, provides a computational resource for solving new problems. We can query the resulting FIP sub-manifold with unseen data by using perplexity (Supplementary) as a measure of a networks intrinsic ability to separate an unseen data set. Using perplexity, we queried the FIP sub-manifold of BERT networks with IMDB data and Wikitext data. We found that the distinct language data sets achieve minimal perplexity on different classes of networks. Wikitext obtains optimal performance on CoCL networks while IMDB achieves optimal performance on CLCo networks. These results demonstrate that the FIP framework can generate diverse sets of neural networks by transforming networks using distinct meta-tasks. The sub-manifolds of weight space can, then, be queried inexpensively to identify networks pre-optimized for new machine learning problems.

\subsection*{Related work}

The major contribution of our framework is that it provides a unified theoretical and mathematical framework through which a set of problems can be addressed across a variety of neural network architectures. To our knowledge, no single mathematical framework has been applied to address the set of machine learning meta problems and diversity of neural network architecture that we address with the functionally invariant path framework. Further, with the rise of transformer models, we have seen the emergence of architecture specific strategies for sparsification and continual learning. We provide a unified framework that can achieve similar quality of results across different classes of architectures (CNN, transformers) and across different transformer variants. While representation learning frameworks have been pursued in fields including Reinforcement Learning \cite{lyle2021effect,jaderberg2016reinforcement}, our work develops a geometric framework that identifies invariant sub-spaces within parameter space for distinct auxiliary tasks. We demonstrate that the framework scales to transformer models with hundreds of millions of parameters. Therefore, our framework provides a unified theoretical and mathematical for connecting the geometry of parameter space with different model sub-tasks while also practically scaling to provide results in line with state of the art, more specific approaches. Here, we review some of the specific approaches that have been studied for continual learning, sparsification, and adversarial robustness. 

\subsubsection*{Continual learning and catastrophic forgetting:} 
Continual learning and catastrophic forgetting have been extensively studied in the machine learning literature. Continual learning (CL) has been studied extensively in the machine learning literature \cite{chen2018lifelong,parisi2019continual,alzubaidi2021review} for classical CNN as well as vision and language transformers \cite{ramasesh2022effect,mirzadeh2022architecture}. A wide variety of approaches have been developed to train neural networks on a sequence of tasks without loss of prior task performance including regularization based methods, weight parameter isolation methods, and replay based methods. Many classic CL/CF approaches were developed for CNN architectures with $<100M$ parameters. With the emergence of transformer models for natural language processing and computer vision, approaches have become more model specific with approaches such as 
Regularization based methods and parameter isolation methods:  Regularization-based methods assigning constraints to prevent parameter changes to important parameters from prior tasks. The functionally invariant paths framework can be viewed as a generalized regularization framework that applies at any point in parameter space. Methods like Elastic Weight Consolidation \cite{kirkpatrick2017overcoming}, orthogonal gradient descent (OGD)\cite{farajtabar2020orthogonal} and Synaptic Intelligence \cite{zenke2017continual} penalize the movement of parameters that are important for solving previous tasks in order to mitigate catastrophic forgetting. As a specific example OGD constrains movement in parameter space to move in a subspace that is orthogonal to gradients from prior tasks. OGD was applied to continual learning tasks on the MNIST data in \cite{farajtabar2020orthogonal} on a small, three layer multi-layer perceptron. The OGD method requires knowledge of prior task gradients and is a local method, like EWC. While locally in parameter space, it makes sense to constrain changes in network parameters along the orthogonal subspace, there is no mathematical reason that long range updates should satisfy this constrain. Past gradient directions become less meaningful as we move away from an initial trained network and begin traversing long range paths in parameter space in search of good networks. By using a metric tensor construction that is defined at every point in parameter space, the FIP framework can traverse long range paths in parameter space sometimes making weight updates that are, in fact, not orthogonal to gradients. 

Replay-based methods store and replay past data or apply generative models to replay data during training to mitigate catastrophic forgetting \cite{rolnick2019experience,shin2017continual}. Regularization-based methods including elastic weight consolidation (EWC) constrain the learning of important parameters from previous tasks by assigning constraints to prevent parameter changes. Architecture expansion or dynamic architecture methods like PNN \cite{rusu2016progressive} and Dynamically Expandable Networks  DEN \cite{yoon2017lifelong} and super masks in position \cite{wortsman2020supermasks} adapt, grow, or mask the model's architecture allowing specialization without interfering with previous knowledge. Parameter isolation methods: These methods isolate specific parameters or subsets of parameters to be task-specific, preventing interference with previous knowledge. Examples include the Context-Dependent Gating (CDG) approach or the Learning without Forgetting (LwF) method.Task-specific model components Instead of modifying the learning objective or replaying data, various methods [42, 53, 31, 30, 32, 52, 4, 11, 51] use different model components for different tasks. In Progressive Neural Networks (PNN), Dynamically Expandable Networks (DEN), and Reinforced Continual Learning (RCL) [42, 53, 52], the model is expanded for each new task. Adapter-BERT \cite{houlsby2019parameter} and Vision transformer (ViT) meta-attention \cite{xue2022meta}. 

\subsubsection*{Fine-tuning and LORA:} LORA 
LoRA (low-rank) 
While several works (including now famously LoRA) have pointed to the potential use of invariant subspaces for model adaptation, we are not aware of mathematical definitions analogous to our metric tensor construction that enable these directions to be defined mathematically or discovered algorithmically. LoRA, for example, to the best of our knowledge introduces a low-rank structure heuristic into the weight update through matrix factorization forcing delta W = A B where A and B have an inner dimension r, and thus controlling the rank of W. FIP is compatible with LoRA adjusted strategies, and in future work we seek to incorporate low-rank updates into the FIP search algorithim. 

\subsubsection*{Sparsification methods:}
Sparsity methods for neural networks aim to reduce the number of parameters or activations in a network, thereby improving efficiency and reducing memory requirements. Unstructured pruning techniques apply strategies based on network weights or weight gradients to remove weights that do not contribute to network performance. Unstructured sparsity methods seek to remove weights at any position in the network. The lottery ticket hypothesis (LTH) demonstrated that dense networks often contain sparse sub-networks, named winning tickets, that can achieve high accuracy when isolated from the dense network  \cite{frankle2018lottery}.  LTH methods discover these sparse networks through empirical pruning procedure. Unstructured pruning methods remove insignificant weight elements using criteria including  weight magnitude, gradient and hessian \cite{han2015learning,alzubaidi2021review,lecun1989optimal}. Recent strategies \cite{evci2020rigging,liu2021we} dynamically extract and train sparse subnetworks instead of training the full models. Evolutionary strategies (SET)\cite{mocanu2018scalable,liu2021sparse} begin with sparse topologies (eg Erdős–Rényi generated graphs) in training and optimize topology and network weights during training. Historically, sparsity methods have been applied to convolutional and multi-layer perceptron architectures. Recent frameworks have been introduced for sparsification of transformer architectures. For example, \cite{chen2021chasing} developed a vision transformer sparsification strategy SViTE that uses integrates model training with prune/grow strategies achieving sparsification of the ViT family transformers.  \cite{brahma2022breaking} explores sparsification of BERT by identify specific internal network topologies that can achieve high performance on natural language processing tasks with sparse weight distribution through empirical investigation and ablation experiments. 

\subsubsection*{Robustness to Adversarial Attack}
Finally, we demonstrate that the FIP can generate ensembles of networks with good performance in adversarial attack paradigms. Like continual learning and sparsificaiton, Adversarial attack mitigation has been addressed by a wide range of methods including augmented training with adversarial examples \cite{madry2017towards}, through the use of diversified network ensembles \cite{pang2019improving}, as well as through re-coding of the input \cite{buckman2018thermometer}. We demonstrate that the FIP framework can generate network ensembles that have intrinsic resistance to adversarial attack when compared with base networks. 

\subsection*{Discussion}

We have introduced a mathematical theory and algorithm for training path connected sets of neural networks to solve machine learning problems.  We demonstrate that path connected sets of networks can be applied to diversify the functional behavior of a network, enabling the network to accommodate additional tasks, to prune weights, or generate diverse ensembles of networks for preventing failure to adversarial attack. More broadly, our mathematical framework provides a useful conceptual view of neural network training. We view a network as a mathematical object that can be transformed through iterative application of distinct meta operations. Meta-operations move the network along paths within the weight space of a neural network. Thus, we identify path connected sub-manifolds of weight space that are specialized for different goals. These sub-manifolds can be enumerated using the FIP algorithm and then queried as a computational resource and applied to solve new problems (Fig. \ref{fig:fig5}E). 

Fundamentally, our work exploits a parameter degeneracy that is intrinsic to large mathematical models. Previous work has demonstrated that large statistical models often contain significant parameter degeneracy \cite{machta2013parameter} leading to `flat' objective functions \cite{hochreiter1997flat,hochreiter1994simplifying,tsuzuku2020normalized}. Recent work in physics has demonstrated that physical models with large numbers of parameters often contain parameter degeneracy such that model parameters can be set to any value within a sub-manifold of parameter space without loss of accuracy in predicting experimental data \cite{machta2013parameter}. Mathematically, in the supporting materials we show that the neural networks that we analyze have mathematical signatures of parameter degeneracy after training, through spectral analysis of the metric tensor.  Modern deep neural networks contain large numbers of parameters that are fit based on training data, and so are related mathematical objects to physical models with large numbers of parameters set by experimental data , and in fact, exact mappings between statistical mechanics models and neural networks exist \cite{mehta2014exact}. Spectral analysis demonstrates that the weight space contains significant sub-spaces  where movement of parameters causes insignificant change in network behavior. Our FIP algorithm explores these degenerate sub-spaces or sub-manifolds of parameter space. Implicitly we show that exploration of the degenerate sub-space can find regions of flexibility where parameters can accommodate a second task (a second image classification task) or goal like sparsification. We apply basic methods from differential geometry to identify and traverse these degenerate sub-spaces. In the Supporting Materials we show that additional concepts from differential geometry including the covariant derivative along a weight space path can be applied to refine paths by minimizing not only the velocity along a weight space path but also acceleration.

Broadly, our results shift attention from the study of single networks to the path-connected sets of neural networks.  Biological systems have been hypothesized to explore a range of effective network configurations due to both fluctuation induced drift and modulation of a given network by other sub-systems within the brain \cite{mau2020brain, stringer2019spontaneous, masset2022drifting,geva2023time,driscoll2022representational}. By shifting attention from networks as single configurations or points in weight space to exploring sub-manifolds of the weight space, the FIP framework may help illuminate a potential principle of flexible intelligence and motivate the development of mathematical methods for studying the local and global geometry of functionally invariant solution sets to machine learning problems \cite{smale1998mathematical,mumford2010pattern}. 


\subsection*{Data Availability}
All data sets used in the manuscript were obtained from public data sources and repositories.  A complete list of public data sources with links can be found at \url{https://github.com/Guruprasad93/FlexibleMachineLearning/blob/main/README.md}. 

\subsection*{Code Availability}
Executable Python code and documentation applying the FIP path finding algorithms to example problems can be found at \url{https://github.com/Guruprasad93/FlexibleMachineLearning}.The code is provided without restriction under the MIT license. 

\bibliography{references,scibib}

\begin{thebibliography}{10}

\bibitem{alzubaidi2021review}
Laith Alzubaidi, Jinglan Zhang, Amjad~J Humaidi, Ayad Al-Dujaili, Ye~Duan,
  Omran Al-Shamma, Jos{\'e} Santamar{\'\i}a, Mohammed~A Fadhel, Muthana
  Al-Amidie, and Laith Farhan.
\newblock Review of deep learning: Concepts, cnn architectures, challenges,
  applications, future directions.
\newblock {\em Journal of big Data}, 8:1--74, 2021.

\bibitem{amari2016information}
Shun-ichi Amari.
\newblock {\em Information geometry and its applications}, volume 194.
\newblock Springer, 2016.

\bibitem{benn1987introduction}
Ian Benn and Robin Tucker.
\newblock {\em An introduction to spinors and geometry with applications in
  physics}.
\newblock Adam Hilger Ltd, 1987.

\bibitem{blalock2020state}
Davis Blalock, Jose Javier~Gonzalez Ortiz, Jonathan Frankle, and John Guttag.
\newblock What is the state of neural network pruning?
\newblock {\em arXiv preprint arXiv:2003.03033}, 2020.

\bibitem{bommasani2021opportunities}
Rishi Bommasani, Drew~A Hudson, Ehsan Adeli, Russ Altman, Simran Arora, Sydney
  von Arx, Michael~S Bernstein, Jeannette Bohg, Antoine Bosselut, Emma
  Brunskill, et~al.
\newblock On the opportunities and risks of foundation models.
\newblock {\em arXiv preprint arXiv:2108.07258}, 2021.

\bibitem{brahma2022breaking}
Siddhartha Brahma, Polina Zablotskaia, and David Mimno.
\newblock Breaking bert: Evaluating and optimizing sparsified attention.
\newblock {\em arXiv preprint arXiv:2210.03841}, 2022.

\bibitem{brown2020language}
Tom Brown, Benjamin Mann, Nick Ryder, Melanie Subbiah, Jared~D Kaplan, Prafulla
  Dhariwal, Arvind Neelakantan, Pranav Shyam, Girish Sastry, Amanda Askell,
  et~al.
\newblock Language models are few-shot learners.
\newblock {\em Advances in neural information processing systems},
  33:1877--1901, 2020.

\bibitem{buckman2018thermometer}
Jacob Buckman, Aurko Roy, Colin Raffel, and Ian Goodfellow.
\newblock Thermometer encoding: One hot way to resist adversarial examples.
\newblock In {\em International conference on learning representations}, 2018.

\bibitem{chen2021chasing}
Tianlong Chen, Yu~Cheng, Zhe Gan, Lu~Yuan, Lei Zhang, and Zhangyang Wang.
\newblock Chasing sparsity in vision transformers: An end-to-end exploration.
\newblock {\em Advances in Neural Information Processing Systems},
  34:19974--19988, 2021.

\bibitem{chen2018lifelong}
Zhiyuan Chen and Bing Liu.
\newblock Lifelong machine learning.
\newblock {\em Synthesis Lectures on Artificial Intelligence and Machine
  Learning}, 12(3):1--207, 2018.

\bibitem{devlin2018bert}
Jacob Devlin, Ming-Wei Chang, Kenton Lee, and Kristina Toutanova.
\newblock Bert: Pre-training of deep bidirectional transformers for language
  understanding.
\newblock {\em arXiv preprint arXiv:1810.04805}, 2018.

\bibitem{dosovitskiy2020image}
Alexey Dosovitskiy, Lucas Beyer, Alexander Kolesnikov, Dirk Weissenborn,
  Xiaohua Zhai, Thomas Unterthiner, Mostafa Dehghani, Matthias Minderer, Georg
  Heigold, Sylvain Gelly, et~al.
\newblock An image is worth 16x16 words: Transformers for image recognition at
  scale.
\newblock {\em arXiv preprint arXiv:2010.11929}, 2020.

\bibitem{driscoll2022representational}
Laura~N Driscoll, Lea Duncker, and Christopher~D Harvey.
\newblock Representational drift: Emerging theories for continual learning and
  experimental future directions.
\newblock {\em Current Opinion in Neurobiology}, 76:102609, 2022.

\bibitem{evci2020rigging}
Utku Evci, Trevor Gale, Jacob Menick, Pablo~Samuel Castro, and Erich Elsen.
\newblock Rigging the lottery: Making all tickets winners.
\newblock In {\em International Conference on Machine Learning}, pages
  2943--2952. PMLR, 2020.

\bibitem{farajtabar2020orthogonal}
Mehrdad Farajtabar, Navid Azizan, Alex Mott, and Ang Li.
\newblock Orthogonal gradient descent for continual learning.
\newblock In {\em International Conference on Artificial Intelligence and
  Statistics}, pages 3762--3773. PMLR, 2020.

\bibitem{frankle2018lottery}
Jonathan Frankle and Michael Carbin.
\newblock The lottery ticket hypothesis: Finding sparse, trainable neural
  networks.
\newblock {\em arXiv preprint arXiv:1803.03635}, 2018.

\bibitem{geva2023time}
Nitzan Geva, Daniel Deitch, Alon Rubin, and Yaniv Ziv.
\newblock Time and experience differentially affect distinct aspects of
  hippocampal representational drift.
\newblock {\em Neuron}, 2023.

\bibitem{han2015learning}
Song Han, Jeff Pool, John Tran, and William Dally.
\newblock Learning both weights and connections for efficient neural network.
\newblock {\em Advances in neural information processing systems}, 28, 2015.

\bibitem{he2022masked}
Kaiming He, Xinlei Chen, Saining Xie, Yanghao Li, Piotr Doll{\'a}r, and Ross
  Girshick.
\newblock Masked autoencoders are scalable vision learners.
\newblock In {\em Proceedings of the IEEE/CVF Conference on Computer Vision and
  Pattern Recognition}, pages 16000--16009, 2022.

\bibitem{hochreiter1994simplifying}
Sepp Hochreiter and J{\"u}rgen Schmidhuber.
\newblock Simplifying neural nets by discovering flat minima.
\newblock {\em Advances in neural information processing systems}, 7, 1994.

\bibitem{hochreiter1997flat}
Sepp Hochreiter and J{\"u}rgen Schmidhuber.
\newblock Flat minima.
\newblock {\em Neural computation}, 9(1):1--42, 1997.

\bibitem{hoffmann2022empirical}
Jordan Hoffmann, Sebastian Borgeaud, Arthur Mensch, Elena Buchatskaya, Trevor
  Cai, Eliza Rutherford, Diego de~Las~Casas, Lisa~Anne Hendricks, Johannes
  Welbl, Aidan Clark, et~al.
\newblock An empirical analysis of compute-optimal large language model
  training.
\newblock {\em Advances in Neural Information Processing Systems},
  35:30016--30030, 2022.

\bibitem{houlsby2019parameter}
Neil Houlsby, Andrei Giurgiu, Stanislaw Jastrzebski, Bruna Morrone, Quentin
  De~Laroussilhe, Andrea Gesmundo, Mona Attariyan, and Sylvain Gelly.
\newblock Parameter-efficient transfer learning for nlp.
\newblock In {\em International Conference on Machine Learning}, pages
  2790--2799. PMLR, 2019.

\bibitem{hu2021lora}
Edward~J Hu, Yelong Shen, Phillip Wallis, Zeyuan Allen-Zhu, Yuanzhi Li, Shean
  Wang, Lu~Wang, and Weizhu Chen.
\newblock Lora: Low-rank adaptation of large language models.
\newblock {\em arXiv preprint arXiv:2106.09685}, 2021.

\bibitem{jaderberg2016reinforcement}
Max Jaderberg, Volodymyr Mnih, Wojciech~Marian Czarnecki, Tom Schaul, Joel~Z
  Leibo, David Silver, and Koray Kavukcuoglu.
\newblock Reinforcement learning with unsupervised auxiliary tasks.
\newblock {\em arXiv preprint arXiv:1611.05397}, 2016.

\bibitem{kaushik2021understanding}
Prakhar Kaushik, Alex Gain, Adam Kortylewski, and Alan Yuille.
\newblock Understanding catastrophic forgetting and remembering in continual
  learning with optimal relevance mapping.
\newblock {\em arXiv preprint arXiv:2102.11343}, 2021.

\bibitem{kirkpatrick2017overcoming}
James Kirkpatrick, Razvan Pascanu, Neil Rabinowitz, Joel Veness, Guillaume
  Desjardins, Andrei~A Rusu, Kieran Milan, John Quan, Tiago Ramalho, Agnieszka
  Grabska-Barwinska, et~al.
\newblock Overcoming catastrophic forgetting in neural networks.
\newblock {\em Proceedings of the national academy of sciences},
  114(13):3521--3526, 2017.

\bibitem{lecun1989optimal}
Yann LeCun, John Denker, and Sara Solla.
\newblock Optimal brain damage.
\newblock {\em Advances in neural information processing systems}, 2, 1989.

\bibitem{liu2021sparse}
Shiwei Liu, Decebal~Constantin Mocanu, Amarsagar Reddy~Ramapuram Matavalam,
  Yulong Pei, and Mykola Pechenizkiy.
\newblock Sparse evolutionary deep learning with over one million artificial
  neurons on commodity hardware.
\newblock {\em Neural Computing and Applications}, 33:2589--2604, 2021.

\bibitem{liu2021we}
Shiwei Liu, Lu~Yin, Decebal~Constantin Mocanu, and Mykola Pechenizkiy.
\newblock Do we actually need dense over-parameterization? in-time
  over-parameterization in sparse training.
\newblock In {\em International Conference on Machine Learning}, pages
  6989--7000. PMLR, 2021.

\bibitem{liu2019roberta}
Yinhan Liu, Myle Ott, Naman Goyal, Jingfei Du, Mandar Joshi, Danqi Chen, Omer
  Levy, Mike Lewis, Luke Zettlemoyer, and Veselin Stoyanov.
\newblock Roberta: A robustly optimized bert pretraining approach.
\newblock {\em arXiv preprint arXiv:1907.11692}, 2019.

\bibitem{lyle2021effect}
Clare Lyle, Mark Rowland, Georg Ostrovski, and Will Dabney.
\newblock On the effect of auxiliary tasks on representation dynamics.
\newblock In {\em International Conference on Artificial Intelligence and
  Statistics}, pages 1--9. PMLR, 2021.

\bibitem{mache2006trends}
Detlef~H Mache, J{\'o}zsef Szabados, and Marcel~G de~Bruin.
\newblock {\em Trends and Applications in Constructive Approximation}, volume
  151.
\newblock Springer Science \& Business Media, 2006.

\bibitem{machta2013parameter}
Benjamin~B Machta, Ricky Chachra, Mark~K Transtrum, and James~P Sethna.
\newblock Parameter space compression underlies emergent theories and
  predictive models.
\newblock {\em Science}, 342(6158):604--607, 2013.

\bibitem{madry2017towards}
Aleksander Madry, Aleksandar Makelov, Ludwig Schmidt, Dimitris Tsipras, and
  Adrian Vladu.
\newblock Towards deep learning models resistant to adversarial attacks.
\newblock {\em arXiv preprint arXiv:1706.06083}, 2017.

\bibitem{masset2022drifting}
Paul Masset, Shanshan Qin, and Jacob~A Zavatone-Veth.
\newblock Drifting neuronal representations: Bug or feature?
\newblock {\em Biological cybernetics}, 116(3):253--266, 2022.

\bibitem{mau2020brain}
William Mau, Michael~E Hasselmo, and Denise~J Cai.
\newblock The brain in motion: How ensemble fluidity drives memory-updating and
  flexibility.
\newblock {\em Elife}, 9:e63550, 2020.

\bibitem{mehta2014exact}
Pankaj Mehta and David~J Schwab.
\newblock An exact mapping between the variational renormalization group and
  deep learning.
\newblock {\em arXiv preprint arXiv:1410.3831}, 2014.

\bibitem{minxha2020flexible}
Juri Minxha, Ralph Adolphs, Stefano Fusi, Adam~N Mamelak, and Ueli Rutishauser.
\newblock Flexible recruitment of memory-based choice representations by the
  human medial frontal cortex.
\newblock {\em Science}, 368(6498):eaba3313, 2020.

\bibitem{mirzadeh2022architecture}
Seyed~Iman Mirzadeh, Arslan Chaudhry, Dong Yin, Timothy Nguyen, Razvan Pascanu,
  Dilan Gorur, and Mehrdad Farajtabar.
\newblock Architecture matters in continual learning.
\newblock {\em arXiv preprint arXiv:2202.00275}, 2022.

\bibitem{mocanu2018scalable}
Decebal~Constantin Mocanu, Elena Mocanu, Peter Stone, Phuong~H Nguyen,
  Madeleine Gibescu, and Antonio Liotta.
\newblock Scalable training of artificial neural networks with adaptive sparse
  connectivity inspired by network science.
\newblock {\em Nature communications}, 9(1):2383, 2018.

\bibitem{mumford2010pattern}
David Mumford and Agn{\`e}s Desolneux.
\newblock {\em Pattern theory: the stochastic analysis of real-world signals}.
\newblock CRC Press, 2010.

\bibitem{openai2023gpt4}
OpenAI.
\newblock Gpt-4 technical report, 2023.

\bibitem{pang2019improving}
Tianyu Pang, Kun Xu, Chao Du, Ning Chen, and Jun Zhu.
\newblock Improving adversarial robustness via promoting ensemble diversity.
\newblock In {\em International Conference on Machine Learning}, pages
  4970--4979. PMLR, 2019.

\bibitem{parisi2019continual}
German~I Parisi, Ronald Kemker, Jose~L Part, Christopher Kanan, and Stefan
  Wermter.
\newblock Continual lifelong learning with neural networks: A review.
\newblock {\em Neural networks}, 113:54--71, 2019.

\bibitem{ramasesh2022effect}
Vinay~Venkatesh Ramasesh, Aitor Lewkowycz, and Ethan Dyer.
\newblock Effect of scale on catastrophic forgetting in neural networks.
\newblock In {\em International Conference on Learning Representations}, 2022.

\bibitem{ramesh2021model}
Rahul Ramesh and Pratik Chaudhari.
\newblock Model zoo: A growing" brain" that learns continually.
\newblock {\em arXiv preprint arXiv:2106.03027}, 2021.

\bibitem{rolnick2019experience}
David Rolnick, Arun Ahuja, Jonathan Schwarz, Timothy Lillicrap, and Gregory
  Wayne.
\newblock Experience replay for continual learning.
\newblock {\em Advances in Neural Information Processing Systems}, 32, 2019.

\bibitem{rumelhart1986learning}
David~E Rumelhart, Geoffrey~E Hinton, and Ronald~J Williams.
\newblock Learning representations by back-propagating errors.
\newblock {\em nature}, 323(6088):533--536, 1986.

\bibitem{rusu2016progressive}
Andrei~A Rusu, Neil~C Rabinowitz, Guillaume Desjardins, Hubert Soyer, James
  Kirkpatrick, Koray Kavukcuoglu, Razvan Pascanu, and Raia Hadsell.
\newblock Progressive neural networks.
\newblock {\em arXiv preprint arXiv:1606.04671}, 2016.

\bibitem{shin2017continual}
Hanul Shin, Jung~Kwon Lee, Jaehong Kim, and Jiwon Kim.
\newblock Continual learning with deep generative replay.
\newblock {\em Advances in neural information processing systems}, 30, 2017.

\bibitem{smale1998mathematical}
Steve Smale.
\newblock Mathematical problems for the next century.
\newblock {\em The mathematical intelligencer}, 20(2):7--15, 1998.

\bibitem{stringer2019spontaneous}
Carsen Stringer, Marius Pachitariu, Nicholas Steinmetz, Charu~Bai Reddy, Matteo
  Carandini, and Kenneth~D Harris.
\newblock Spontaneous behaviors drive multidimensional, brainwide activity.
\newblock {\em Science}, 364(6437):eaav7893, 2019.

\bibitem{taori2023alpaca}
Rohan Taori, Ishaan Gulrajani, Tianyi Zhang, Yann Dubois, Xuechen Li, Carlos
  Guestrin, Percy Liang, and Tatsunori~B Hashimoto.
\newblock Alpaca: A strong, replicable instruction-following model.
\newblock {\em Stanford Center for Research on Foundation Models. https://crfm.
  stanford. edu/2023/03/13/alpaca. html}, 3(6):7, 2023.

\bibitem{touvron2021training}
Hugo Touvron, Matthieu Cord, Matthijs Douze, Francisco Massa, Alexandre
  Sablayrolles, and Herv{\'e} J{\'e}gou.
\newblock Training data-efficient image transformers \& distillation through
  attention.
\newblock In {\em International conference on machine learning}, pages
  10347--10357. PMLR, 2021.

\bibitem{tsuzuku2020normalized}
Yusuke Tsuzuku, Issei Sato, and Masashi Sugiyama.
\newblock Normalized flat minima: Exploring scale invariant definition of flat
  minima for neural networks using pac-bayesian analysis.
\newblock In {\em International Conference on Machine Learning}, pages
  9636--9647. PMLR, 2020.

\bibitem{van2020brain}
Gido~M van~de Ven, Hava~T Siegelmann, and Andreas~S Tolias.
\newblock Brain-inspired replay for continual learning with artificial neural
  networks.
\newblock {\em Nature communications}, 11(1):1--14, 2020.

\bibitem{vaswani2017attention}
Ashish Vaswani, Noam Shazeer, Niki Parmar, Jakob Uszkoreit, Llion Jones,
  Aidan~N Gomez, {\L}ukasz Kaiser, and Illia Polosukhin.
\newblock Attention is all you need.
\newblock {\em Advances in neural information processing systems}, 30, 2017.

\bibitem{weisstein2014metric}
Eric~W Weisstein.
\newblock Metric tensor.
\newblock {\em https://mathworld. wolfram. com/}, 2014.

\bibitem{wolf2020transformers}
Thomas Wolf, Lysandre Debut, Victor Sanh, Julien Chaumond, Clement Delangue,
  Anthony Moi, Pierric Cistac, Tim Rault, R{\'e}mi Louf, Morgan Funtowicz,
  et~al.
\newblock Transformers: State-of-the-art natural language processing.
\newblock In {\em Proceedings of the 2020 conference on empirical methods in
  natural language processing: system demonstrations}, pages 38--45, 2020.

\bibitem{wortsman2020supermasks}
Mitchell Wortsman, Vivek Ramanujan, Rosanne Liu, Aniruddha Kembhavi, Mohammad
  Rastegari, Jason Yosinski, and Ali Farhadi.
\newblock Supermasks in superposition.
\newblock {\em Advances in Neural Information Processing Systems},
  33:15173--15184, 2020.

\bibitem{xue2022meta}
Mengqi Xue, Haofei Zhang, Jie Song, and Mingli Song.
\newblock Meta-attention for vit-backed continual learning.
\newblock In {\em Proceedings of the IEEE/CVF Conference on Computer Vision and
  Pattern Recognition}, pages 150--159, 2022.

\bibitem{yoon2017lifelong}
Jaehong Yoon, Eunho Yang, Jeongtae Lee, and Sung~Ju Hwang.
\newblock Lifelong learning with dynamically expandable networks.
\newblock {\em arXiv preprint arXiv:1708.01547}, 2017.

\bibitem{zenke2017continual}
Friedemann Zenke, Ben Poole, and Surya Ganguli.
\newblock Continual learning through synaptic intelligence.
\newblock In {\em International conference on machine learning}, pages
  3987--3995. PMLR, 2017.

\end{thebibliography}
\bibliographystyle{plain}




\clearpage

\end{document}